\documentclass{SSBAtran}
\usepackage{amsmath}

\usepackage{graphicx}
\usepackage{url}

\begin{document}
%
\title{Towards Controllable Image Generation through Representation-Conditioned Diffusion Models}

\author{\IEEEauthorblockN{%
Nithesh Chandher Karthikeyan\IEEEauthorrefmark{1}, 
Jonas Unger\IEEEauthorrefmark{1}, and 
Gabriel Eilertsen\IEEEauthorrefmark{1}}

\IEEEauthorblockA{\IEEEauthorrefmark{1}Linköping University, Norrköping, Sweden\\
Email: $\{$nithesh.chandher.karthikeyan,  jonas.unger, gabriel.eilertsen$\}$@liu.se}

}

\maketitle

\begin{abstract}

Diffusion models have emerged as powerful tools for high-quality image generation and editing, but guiding these models to produce specific outputs remains a challenge. Conventional approaches rely on conditioning mechanisms, such as text prompts or semantic maps, which require extensively annotated datasets. In this preliminary work, we explore diffusion models conditioned on representations from a pre-trained self-supervised model. The self-conditioning mechanism not only improves the quality of unconditional image generation, but also provides a representation space that can be used to control the generation. We explore this conditioning space by identifying directions of variations, and demonstrate promising properties in terms of smoothness and disentanglement.
\end{abstract}


\section{Introduction}

Diffusion models have transformed the field of generative modeling, offering a powerful alternative to previous models, such as Generative Adversarial Networks (GANs) \cite{goodfellow2020generative}, for high-quality image synthesis and editing. However, GANs provide a well-structured and disentangled latent space, facilitating precise modifications of fine-grained image details while preserving unrelated attributes \cite{shen2020interpreting, abdal2019image2stylegan, harkonen2020ganspace}. In contrast, formulating a comparable latent space in diffusion models remains a challenging research problem. Over the years, various approaches have been explored to address this challenge, including inversion into the text-conditioning space \cite{zhou2023training}, manipulation of activations of the denoising model \cite{kwon2022diffusion, haas2024discovering}, and joint learning of a semantically meaningful latent space \cite{preechakul2022diffusion}.

Recently, representation-conditioning has been demonstrated for the purpose of improving the quality of unconditional generation, by providing guidance through self-supervision from extracted image representations \cite{li2025return}. We see this as a promising direction not only for improving unconditional generation but also for providing a latent space that can be used for controllable generation. However, to our knowledge, representation-conditioned diffusion models have not previously been thoroughly tested for this purpose.
Thus, in this paper we present initial experiments on controlling diffusion models through a pre-trained image representation space. 

\begin{figure}[t!]
\centering
\includegraphics[width=210pt]{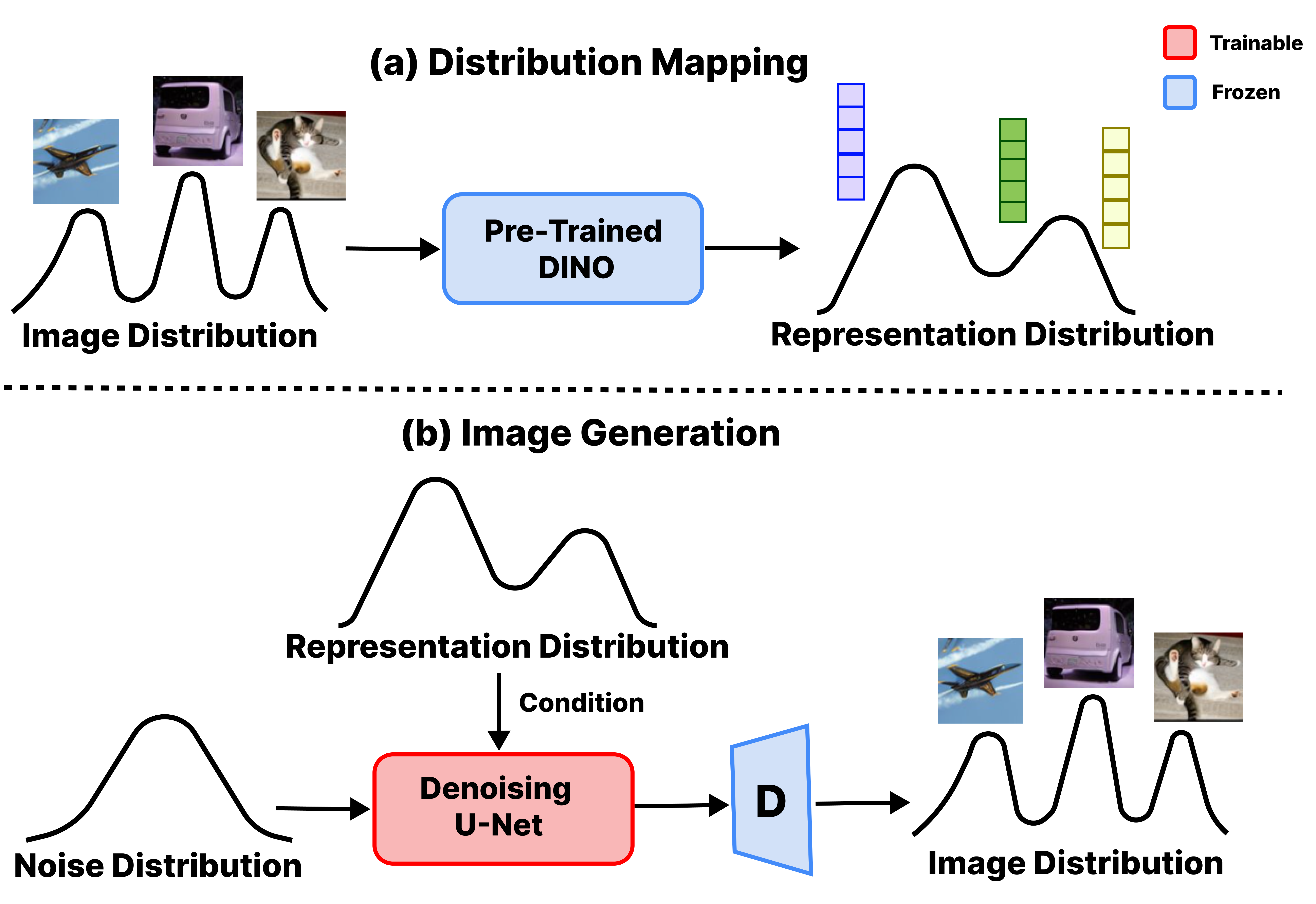}
\caption{The Representation Conditioned Diffusion Model consists of two parts: (a) a pre-trained self-supervised encoder (DINO in our case) maps the image distribution to a representation distribution, and (b) a denoising U-Net maps a noise distribution to the image distribution, conditioned on the representation distribution from (a).}
\label{fig:dino-ldm}
\end{figure}

\section{Related Work}
\textbf{Text Conditioning:} Conventional methods for conditioning diffusion models involve using text prompts, bounding boxes, or semantic maps as input~\cite{rombach2022high}. Textual Inversion \cite{gal2022image} maps a semantic or concept from a few images to a unique word token, whereas DreamBooth \cite{ruiz2023dreambooth} achieves a similar goal by fine-tuning the Imagen model with a few reference images. Custom Diffusion \cite{kumari2023multi} learns multiple concepts by fine-tuning the cross-attention blocks in the diffusion model. However, these methods are highly dependent on effective prompt engineering, which can be limiting when achieving precise control over intricate details and may struggle to adapt to unstructured datasets.

\textbf{H-Space Manipulations:} Kwon et al. \cite{kwon2022diffusion} identifies a latent space based on activations of the denoising model over various timesteps. This semantic latent space, known as H-space, is utilized to discover interpretable and disentangled directions \cite{haas2024discovering}. In contrast, our goal is to establish a well-structured latent space that enables direct sampling and controlled generation.

\begin{figure*}[ht!]
\centering
\includegraphics[width=0.9\textwidth]{images/result-2.jpg}
\vspace{-3mm}
\caption{Image variations are created by linearly interpolating between two embedding vectors, derived from the corresponding reference image or prompt, using different interpolation factors ($\alpha$). The images are generated using three different methods, with the respective reference prompt or image displayed on both ends.}
\label{fig:interpolation}
\end{figure*}

\textbf{Representation Conditioning:}
Early research on Representation Conditioned Diffusion Models (RCDM) investigate how much information from the input image is preserved in the generated output when conditioned on its representation \cite{bordes2021high}. Representation Conditioned Generation (RCG) \cite{li2025return} uses representations from a self-supervised model to condition an image generator. However, it introduces an additional layer by training a representation generator for conditioning, enabling fully unsupervised image generation. Diffusion Autoencoders (Diffusion AEs) \cite{preechakul2022diffusion} employ an explicit semantic encoder that learns some representation from the training images, alongside a conditional DDIM that acts both as a stochastic encoder and decoder. However, unlike RCG, Diffusion AEs do not provide the same quality improvement in unconditional generation.

\section{Method}
\label{method}
Training a Representation Conditioned Diffusion Model (RCDM) can be broken down into two stages: First, images are mapped to a lower-dimensional space using a pre-trained Self-Supervised Learning (SSL) encoder. Then, a diffusion model is trained while conditioning on these representations (see Figure~\ref{fig:dino-ldm}). 
In this work, we adapt the method from \cite{li2025return}, which uses the MoCo v3 encoder and the MAGE generator, to generation through the latent diffusion model (LDM) \cite{rombach2022high} conditioned on DINO representations \cite{caron2021emerging}. That is, denoising in the LDM latent space is guided by the information provided by an image's DINO representation. While it is possible to train a separate representation generator for fully unconditioned generation, in this work we perform experiments on representations from a separate test dataset.

\subsection{Data and Training Setup}
For the experiments, two datasets were used: (a) LSUN Churches \cite{yu2015lsun} and (b) Celeb-A \cite{liu2018large}. The pre-trained DINO encoder \cite{caron2021emerging} extracts 768-dimensional representations from these datasets. For image compression in LDM, a pre-trained VAE model with KL-divergence loss \cite{van2017neural} was used, and RCDMs were trained for each dataset using a conditional de-noising U-Net. 

\section{Properties of the Representation Space}
In this section, we perform experiments to validate the effectiveness of using the representation space for controllable image generation.

\subsection{Perturbation and Interpolation}
\begin{figure}[t]
\centering
\includegraphics[width=240pt]{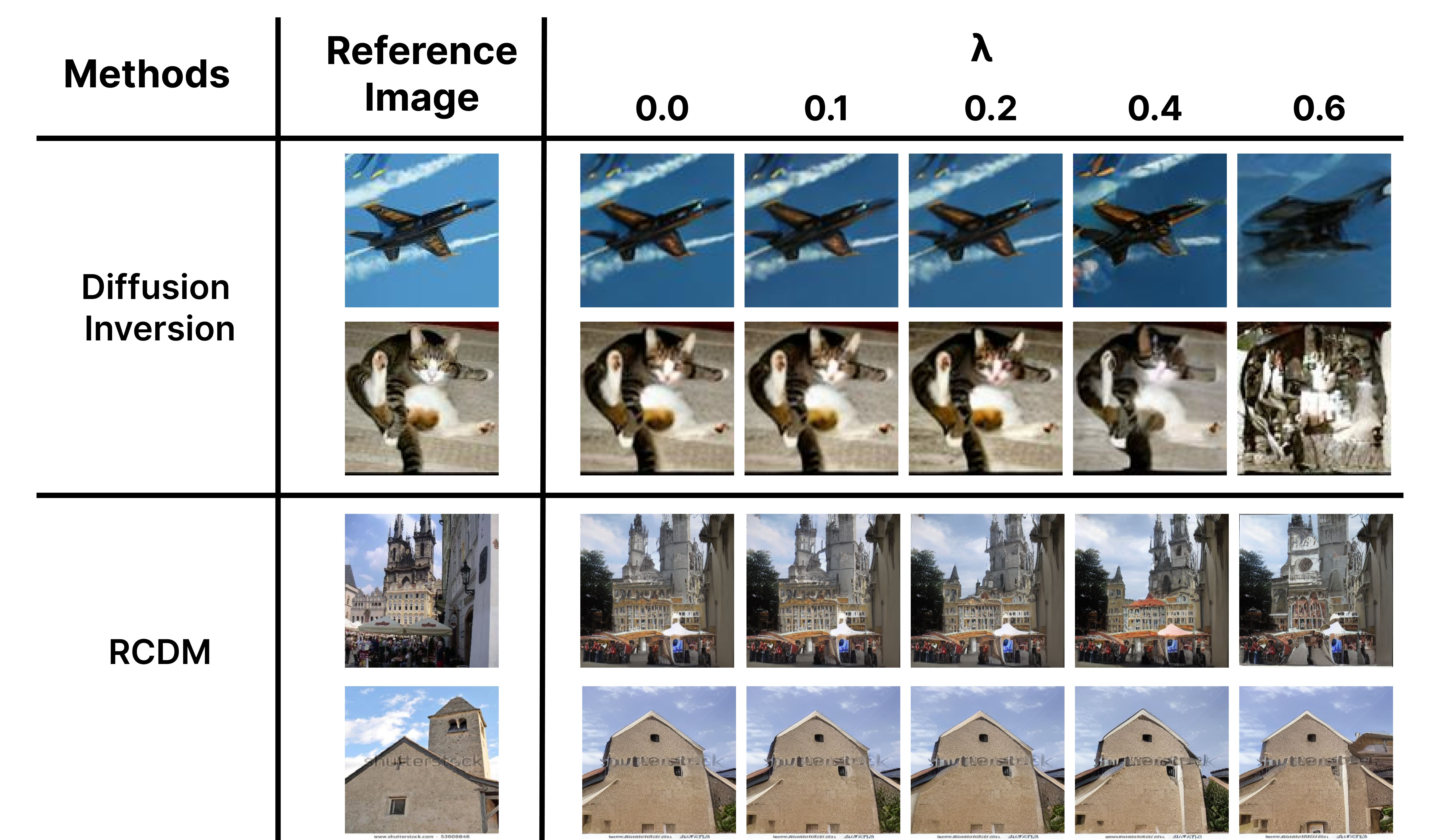}
\vspace{-3mm}
\caption{Images generated by perturbing the embedding vector $\mathcal{\hat{C}}$ (from the reference image) using Gaussian noise with varying strength $\lambda$.}
\label{fig:noise-perturbation}
\end{figure}

Given a representation vector \( \mathcal{C} \), we sample a noise vector \( \epsilon \sim \mathcal{N}(0,1) \) and perturb the representation vector according to $\mathcal{\hat{C}} = \mathcal{C} + \lambda\epsilon$,
where \( \lambda \) controls the perturbation strength. Figure \ref{fig:noise-perturbation} shows the generated images at different perturbation strengths for two methods: Diffusion Inversion \cite{zhou2023training}, trained on the STL-10 dataset \cite{coates2011analysis} and RCDM, trained on LSUN-Churches \cite{yu2015lsun}. At higher noise levels ($\lambda > 0.4$), Diffusion Inversion exhibits significant quality degradation, while RCDM modifies image content while maintaining image quality and consistency. 

We can also perform interpolations in the representation space. Given two representation vectors \( \mathcal{C}_1 \) and \( \mathcal{C}_2 \), a new vector can be obtained by $\mathcal{\hat{C}} = \alpha\mathcal{C}_1 + (1-\alpha)\mathcal{C}_2$.
The resulting image generated from \(\mathcal{\hat{C}}\) will exhibit features of both input images, with the influence of each determined by the interpolation factor \(\alpha\). Figure \ref{fig:interpolation} shows the images generated at different interpolation strengths for three methods. In Stable Diffusion \cite{rombach2022high}, two text embeddings are extracted from separate text prompts using CLIP, and the generated images represent interpolations between these embeddings. For Diffusion Inversion and RCDM, images are produced by interpolating between two learned embeddings or representations (in the case of RCDM).
The results indicate that RCDM provides a semantically smoother transition between reference images. In Diffusion Inversion the images become indistinguishable in the middle, while for Stable Diffusion a sudden transition between images is observed for $\alpha$ values 0.5-0.6. 


\begin{figure*}[ht!]
\centering
\includegraphics[width=0.9\textwidth]{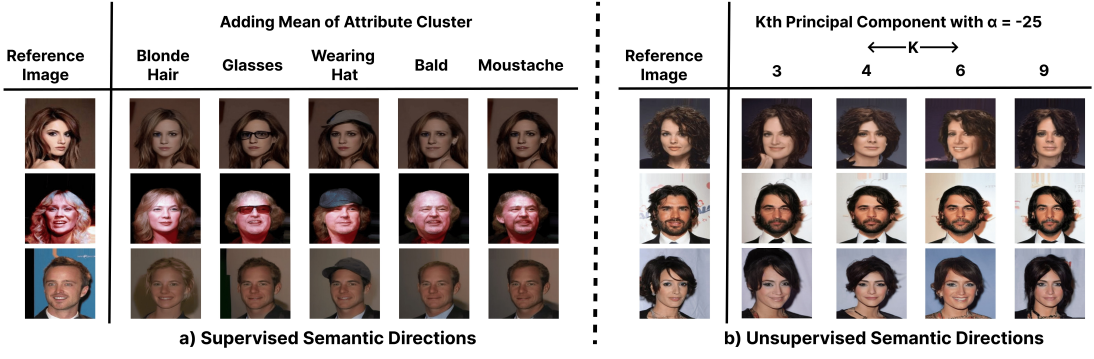}
\vspace{-3mm}
\caption{Images generated by identifying semantic directions within the representation space of diffusion models. (a) supervised approach using the mean representation of a given attribute, (b) unsupervised approach which uses PCA to discover semantic directions.}
\label{fig:pca}
\end{figure*}

\subsection{Semantic Directions}
\paragraph{Supervised Semantic Directions}
In a supervised setting with access to image annotations, identifying semantic directions in the representation space of RCDM becomes more direct. Figure \ref{fig:pca} (a) illustrates examples of discovering meaningful directions in the representation space of RCDM trained on the Celeb-A dataset. To add an attribute like blonde hair (as shown in the figure \ref{fig:pca} (a)) to a reference image, the mean representation of all faces with blonde hair is computed and added to the reference image, enabling the desired feature modification. However, correlations present in the dataset can influence image generation. For example, adding the bald attribute may also unintentionally alter the gender to male, as the dataset contain a higher proportion of bald male images.

\paragraph{Unsupervised Semantic Directions}
Identifying interpretable directions in the representation space becomes challenging in an unsupervised setting. Härkönen et al.~\cite{harkonen2020ganspace} addressed this issue by applying Principal Component Analysis (PCA) to manipulate generated images along meaningful attributes.
We use a similar approach in the representation-conditioning space, perturbing representations according to $\mathcal{\hat{C}} = \mathcal{C} + \alpha V_K$, where $V_K$ is the $K$:th principal component and \( \alpha \) is a scaling factor controlling the strength of the modification.  
Figure \ref{fig:pca} (b) presents the results for RCDM trained on Celeb-A with different components $K$ and a fixed 
$\alpha$ value of $-25$. While these initial results do not show the same amount of disentanglement and interpretability as compared to GANs~\cite{harkonen2020ganspace}, there are variations that suggest it could be possible to utilize unsupervised discovery of directions in the representation space. For example, we can see variations in terms of the size of forehead (K=3), short hair (K=4), background color (K=6), and long hair (K=9).

\section{Conclusion}
In this work, we have demonstrated that representation-conditioned diffusion models exhibit promising properties in the conditioning space, including smoothness and a certain degree of disentanglement. These characteristics suggest that representation conditioning is a compelling approach for improving both the quality and controllability of diffusion models. For future work, representations could be adapted to improve disentanglement, and the results could be used in applications such as controlled generation in unconditional diffusion models for image editing and training data augmentation.



\bibliographystyle{SSBAtrans}
\bibliography{sample}

@article{goodfellow2020generative,
  title={Generative adversarial networks},
  author={Goodfellow, Ian and Pouget-Abadie, Jean and Mirza, Mehdi and Xu, Bing and Warde-Farley, David and Ozair, Sherjil and Courville, Aaron and Bengio, Yoshua},
  journal={Communications of the ACM},
  volume={63},
  number={11},
  pages={139--144},
  year={2020},
  publisher={ACM New York, NY, USA}
}

@inproceedings{shen2020interpreting,
  title={Interpreting the latent space of gans for semantic face editing},
  author={Shen, Yujun and Gu, Jinjin and Tang, Xiaoou and Zhou, Bolei},
  booktitle={Proceedings of the IEEE/CVF conference on computer vision and pattern recognition},
  pages={9243--9252},
  year={2020}
}

@inproceedings{abdal2019image2stylegan,
  title={Image2stylegan: How to embed images into the stylegan latent space?},
  author={Abdal, Rameen and Qin, Yipeng and Wonka, Peter},
  booktitle={Proceedings of the IEEE/CVF international conference on computer vision},
  pages={4432--4441},
  year={2019}
}

@article{harkonen2020ganspace,
  title={Ganspace: Discovering interpretable gan controls},
  author={H{\"a}rk{\"o}nen, Erik and Hertzmann, Aaron and Lehtinen, Jaakko and Paris, Sylvain},
  journal={Advances in neural information processing systems},
  volume={33},
  pages={9841--9850},
  year={2020}
}

@inproceedings{rombach2022high,
  title={High-resolution image synthesis with latent diffusion models},
  author={Rombach, Robin and Blattmann, Andreas and Lorenz, Dominik and Esser, Patrick and Ommer, Bj{\"o}rn},
  booktitle={Proceedings of the IEEE/CVF conference on computer vision and pattern recognition},
  pages={10684--10695},
  year={2022}
}

@inproceedings{ruiz2023dreambooth,
  title={Dreambooth: Fine tuning text-to-image diffusion models for subject-driven generation},
  author={Ruiz, Nataniel and Li, Yuanzhen and Jampani, Varun and Pritch, Yael and Rubinstein, Michael and Aberman, Kfir},
  booktitle={Proceedings of the IEEE/CVF conference on computer vision and pattern recognition},
  pages={22500--22510},
  year={2023}
}

@article{gal2022image,
  title={An image is worth one word: Personalizing text-to-image generation using textual inversion},
  author={Gal, Rinon and Alaluf, Yuval and Atzmon, Yuval and Patashnik, Or and Bermano, Amit H and Chechik, Gal and Cohen-Or, Daniel},
  journal={arXiv preprint arXiv:2208.01618},
  year={2022}
}

@inproceedings{kumari2023multi,
  title={Multi-concept customization of text-to-image diffusion},
  author={Kumari, Nupur and Zhang, Bingliang and Zhang, Richard and Shechtman, Eli and Zhu, Jun-Yan},
  booktitle={Proceedings of the IEEE/CVF Conference on Computer Vision and Pattern Recognition},
  pages={1931--1941},
  year={2023}
}

@inproceedings{caron2021emerging,
  title={Emerging properties in self-supervised vision transformers},
  author={Caron, Mathilde and Touvron, Hugo and Misra, Ishan and J{\'e}gou, Herv{\'e} and Mairal, Julien and Bojanowski, Piotr and Joulin, Armand},
  booktitle={Proceedings of the IEEE/CVF international conference on computer vision},
  pages={9650--9660},
  year={2021}
}

@article{bordes2021high,
  title={High fidelity visualization of what your self-supervised representation knows about},
  author={Bordes, Florian and Balestriero, Randall and Vincent, Pascal},
  journal={arXiv preprint arXiv:2112.09164},
  year={2021}
}

@article{li2025return,
  title={Return of unconditional generation: A self-supervised representation generation method},
  author={Li, Tianhong and Katabi, Dina and He, Kaiming},
  journal={Advances in Neural Information Processing Systems},
  volume={37},
  pages={125441--125468},
  year={2025}
}

@article{van2017neural,
  title={Neural discrete representation learning},
  author={Van Den Oord, Aaron and Vinyals, Oriol and others},
  journal={Advances in neural information processing systems},
  volume={30},
  year={2017}
}

@article{yu2015lsun,
  title={Lsun: Construction of a large-scale image dataset using deep learning with humans in the loop},
  author={Yu, Fisher and Seff, Ari and Zhang, Yinda and Song, Shuran and Funkhouser, Thomas and Xiao, Jianxiong},
  journal={arXiv preprint arXiv:1506.03365},
  year={2015}
}

@article{liu2018large,
  title={Large-scale celebfaces attributes (celeba) dataset},
  author={Liu, Ziwei and Luo, Ping and Wang, Xiaogang and Tang, Xiaoou},
  journal={Retrieved August},
  volume={15},
  number={2018},
  pages={11},
  year={2018}
}

@article{zhou2023training,
  title={Training on thin air: Improve image classification with generated data},
  author={Zhou, Yongchao and Sahak, Hshmat and Ba, Jimmy},
  journal={arXiv preprint arXiv:2305.15316},
  year={2023}
}

@article{kwon2022diffusion,
  title={Diffusion models already have a semantic latent space},
  author={Kwon, Mingi and Jeong, Jaeseok and Uh, Youngjung},
  journal={arXiv preprint arXiv:2210.10960},
  year={2022}
}

@inproceedings{haas2024discovering,
  title={Discovering interpretable directions in the semantic latent space of diffusion models},
  author={Haas, Ren{\'e} and Huberman-Spiegelglas, Inbar and Mulayoff, Rotem and Gra{\ss}hof, Stella and Brandt, Sami S and Michaeli, Tomer},
  booktitle={2024 IEEE 18th International Conference on Automatic Face and Gesture Recognition (FG)},
  pages={1--9},
  year={2024},
  organization={IEEE}
}

@inproceedings{preechakul2022diffusion,
  title={Diffusion autoencoders: Toward a meaningful and decodable representation},
  author={Preechakul, Konpat and Chatthee, Nattanat and Wizadwongsa, Suttisak and Suwajanakorn, Supasorn},
  booktitle={Proceedings of the IEEE/CVF conference on computer vision and pattern recognition},
  pages={10619--10629},
  year={2022}
}

@inproceedings{coates2011analysis,
  title={An analysis of single-layer networks in unsupervised feature learning},
  author={Coates, Adam and Ng, Andrew and Lee, Honglak},
  booktitle={Proceedings of the fourteenth international conference on artificial intelligence and statistics},
  pages={215--223},
  year={2011},
  organization={JMLR Workshop and Conference Proceedings}
}

\end{document}